\documentclass[letterpaper, 10 pt, conference]{ieeeconf}  

\IEEEoverridecommandlockouts                              

\usepackage{graphics} 
\usepackage{epsfig} 
\usepackage{mathptmx} 
\usepackage{times} 
\usepackage{amsmath} 
\usepackage{amssymb}  

\usepackage{graphicx}
\usepackage{booktabs}
\usepackage{makecell}

\usepackage{adjustbox}
\usepackage{multirow}
\usepackage{wrapfig}
\usepackage{url}
\usepackage{hyperref}

\title{\LARGE \bf
Mean Shift Mask Transformer for Unseen Object Instance Segmentation
}

\author{Yangxiao Lu \hspace{3px} Yuqiao Chen \hspace{3px} Nicholas Ruozzi \hspace{3px} Yu Xiang\\
The University of Texas at Dallas \hspace{3px} \\
{\tt\small \{yangxiao.lu, yuqiao.chen, nicholas.ruozzi, yu.xiang\}@utdallas.edu}
}

\author{
Yangxiao Lu, Yuqiao Chen, Nicholas Ruozzi, Yu Xiang
\thanks{Yangxiao Lu, Yuqiao Chen, Nicholas Ruozzi and Yu Xiang are with the Department of Computer Science, the University of Texas at Dallas, Richardson, TX 75080, USA \tt\small \{yangxiao.lu, yuqiao.chen, nicholas.ruozzi, yu.xiang\}@utdallas.edu}
}

\begin{document}

\maketitle
\thispagestyle{empty}
\pagestyle{empty}


\begin{abstract}
Segmenting unseen objects from images is a critical perception skill that a robot needs to acquire. In robot manipulation, it can facilitate a robot to grasp and manipulate unseen objects. Mean shift clustering is a widely used method for image segmentation tasks. However, the traditional mean shift clustering algorithm is not differentiable, making it difficult to integrate it into an end-to-end neural network training framework. In this work, we propose the Mean Shift Mask Transformer (MSMFormer), a new transformer architecture that simulates the von Mises-Fisher (vMF) mean shift clustering algorithm, allowing for the joint training and inference of both the feature extractor and the clustering. Its central component is a hypersphere attention mechanism, which updates object queries on a hypersphere. To illustrate the effectiveness of our method, we apply MSMFormer to unseen object instance segmentation. Our experiments show that MSMFormer achieves competitive performance compared to state-of-the-art methods for unseen object instance segmentation\footnote{Project page, appendix, video, and code are available at: \url{https://irvlutd.github.io/MSMFormer}}.

\end{abstract}

\vspace{-2mm}
\section{INTRODUCTION}
\vspace{-2mm}

The ability to segment unseen objects in cluttered scenes, a problem known as Unseen Object Instance Segmentation (UOIS), is critical in a variety of robotic applications. For example, UOIS allows robots to separate objects from each other to facilitate grasping of these objects~\cite{mousavian20196,sundermeyer2021contact,wang2022goal,goyal2022ifor}.


Unlike category-based object instance segmentation \cite{he2017mask, Carion2020, tian2020conditional, vaswani2017attention, jia2016dynamic, cheng2021per, cheng2022masked}, UOIS requires the ability to generalize to unseen objects, that is, unseen object instances and unseen object categories. To improve generalizability, the state-of-the-art UOIS method~\cite{xiang2020learning} employs a feature learning and clustering strategy, where pixel-wise feature representations are learned using deep neural networks, and then a traditional clustering algorithm, such as mean shift clustering, is utilized to cluster pixel features for object segmentation. Although the feature learning paradigm achieves competitive performance, its potential capabilities are limited by its dependence on traditional clustering algorithms: the cluster assignment steps in these algorithms are not differentiable, which prevents the clustering and feature learning from being combined into a single end-to-end framework. The parameters in the clustering algorithm cannot be learned together with the feature representations.

In this work, we introduce the Mean Shift decoder (MS decoder), a novel differentiable mean shift clustering module based on the transformer architecture~\cite{vaswani2017attention}. Our design of the MS decoder converts iterations of the von Mises-Fisher (vMF) mean shift clustering algorithm into multiple MS decoder layers. At each decoder layer, a set of object queries represent the cluster centers. Cross-attention is applied between the object queries and the pixel features to transform the object queries, which resembles the mean shift procedure to find the cluster centers. In addition, we propose to use L2 normalized object queries and pixel embeddings, and we introduce a hypersphere attention mechanism using the normalized queries, keys, and values. 



\begin{figure}
    \centering
    \includegraphics[width=0.85\columnwidth]{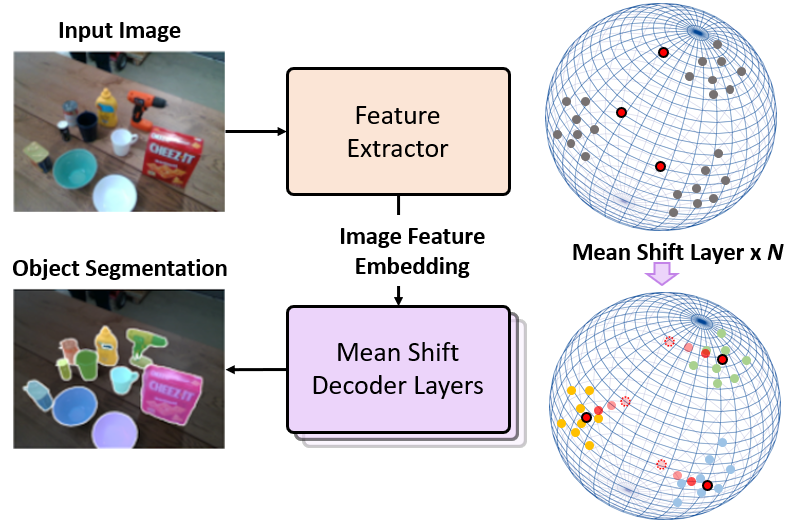}
    \vspace{-2mm}
    \caption{Our framework consists of a feature extractor network to embed image pixels into a high dimensional hypersphere, and the mean shift decoder layers to compute cluster centers of the pixels for unseen object instance segmentation.}
    \label{fig:intro}
    \vspace{-7mm}
\end{figure}

In order to verify the effectiveness of our proposed MS decoder, we introduce a new network, Mean Shift Mask Transformer (MSMFormer), for UOIS by combining a feature learning backbone with the MS decoder for clustering. Since the MS decoder is differentiable, we can train this network end-to-end. Moreover, learning the weights in the MS decoder from data can further increase the capability of the model. Following previous work on UOIS, we train our network on a synthetic RGB-D dataset~\cite{xie2020best} and then test the trained network on the real-world Object Clutter Indoor Dataset (OCID) \cite{suchi2019easylabel} and the Object Segmentation Database (OSD) \cite{richtsfeld2012segmentation}. Our method improves over the state-of-the-art methods on the OCID dataset and achieves comparable performance on the OSD dataset.

Our contributions are summarized as follows. 1) We introduce the mean shift decoder for differentiable mean shift clustering. 2) We introduce a hypersphere attention mechanism to improve training of transformers. 3) We propose MSMFormer for unseen object instance segmentation by combining feature learning and our mean shift decoder.

\section{Related Work}
\vspace{-1mm}

\textbf{Semantic / Instance / Panoptic Segmentation}. These three types of image segmentation tasks are based on object categories. Semantic segmentation labels each pixel in an input image with a class label such as sky, person, table, etc.~\cite{long2015fully,chen2017deeplab}. Instance segmentation requires separating object instances among the same category such as multiple persons or cars. For example, Mask R-CNN \cite{he2017mask} extends Faster R-CNN \cite{Ren2015} by predicting an object mask for each detected object instance. Panoptic segmentation is proposed to unify both semantic and instance segmentation tasks, where every pixel needs to be labeled with a class label and an instance id~\cite{kirillov2019panoptic,cheng2020panoptic,li2022fully}. It is tempting to solve panoptic segmentation tasks to obtain more information from the images. However, as discussed in Mask2Former \cite{cheng2022masked}, panoptic architectures do not guarantee good
performance on instance segmentation tasks since they do not measure the abilities to
rank predictions as instance segmentations. Mask2Former \cite{cheng2022masked} is the first architecture that can achieve competitive performance on all of these segmentation tasks. Hence, we utilize their masked attention~\cite{cheng2022masked} in our method.

\textbf{Unseen Object Instance Segmentation (UOIS)}. UOIS is object category agnostic. The goal is to segment unseen object instances from input images. The general idea for UOIS is to train with a large number of objects and hope the trained model can generalize to unseen objects. Since real images of many objects are difficult to collect, recent techniques utilize synthetic data for training~\cite{shao2018clusternet, xie2020best, danielczuk2019segmenting}. In addition, depth images are widely used in UOIS to improve generalization. Recently, Zhang et al.~\cite{zhang2022unseen} suggest exploiting test-time domain adaption to bridge the gap and boost the segmentation performance. Instead of depth input, Instance Stereo Transformer~\cite{durner2021unknown} generates object masks through the use of stereo image pairs. Unseen Clustering Network (UCN)~\cite{xiang2020learning} employs non-photorealistic RGB images and depth images to learn feature representations to segment unseen objects. It groups pixels via the von Mises-
Fisher (vMF) mean shift clustering algorithm after learning the feature representation of pixels. To maximize image feature performance, our technique replaces the vMF mean shift clustering with a sequence of learnable transformer layers. 

\begin{figure*}
    \vspace{-6mm}
    \centering
\includegraphics[width=1.3\columnwidth]{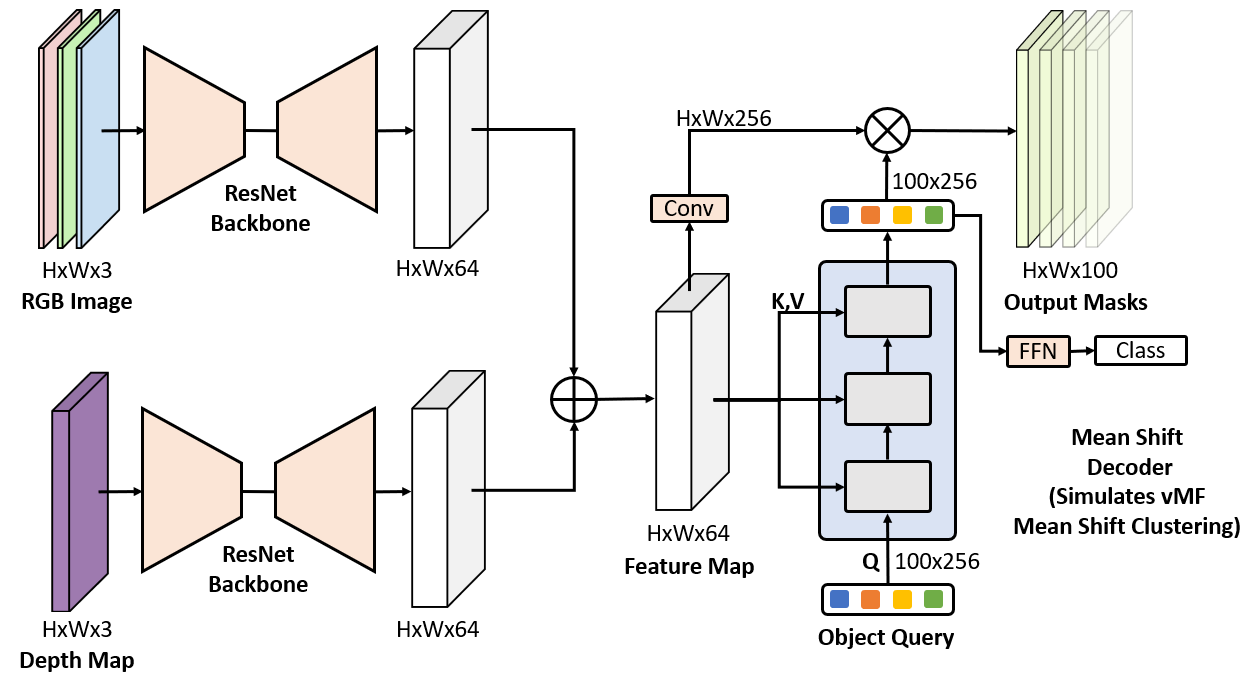}
    \vspace{-4mm}
    \caption{The Mean Shift Mask Transformer consists of backbones and a mean shift decoder. The backbones can be any network that produces pixel embeddings. The depth map input and its backbone are optional depending on the task. A series of mean shift decoder layers are applied to transform $N=100$ object queries. Then these transformed object queries group related pixels to generate masks and predict the classes of each mask. ``FFN'' indicates a Feed-Forward Network. }
    \label{fig:model}
    \vspace{-6mm}
\end{figure*}

\textbf{Differences between K-Means Mask Transformer~\cite{yu2022k} and Ours.}  K-means Mask Xformer \cite{yu2022k} reformulates the cross-attention learning process as a k-means clustering procedure by proposing a new k-means cross-attention. However, the k-means cross-attention is not differentiable due to a hard assignment of pixel features to cluster centers, which makes training difficult. \cite{yu2022k} needs to apply deep supervision after each decoder layer. In contrast, our proposed mean shift cross-attention using hypersphere attention is differentiable. In addition, the K-means Mask Xformer architecture is proposed for panoptic segmentation. Its performance on instance segmentation is not guaranteed. We were unable to successfully train a UOIS model using the K-means Mask Xformer architecture in our experiments.

\textbf{vMF Mean Shift Clustering}. The von Mises-Fisher (vMF) \cite{mardia2000directional} distribution defines a distribution over unit vectors $\mathbf{x}$, $\| \mathbf{x} \|=1$, on the $d$ dimensional hypersphere. The vMF density function is defined as $P(\mathbf{x}; \mathbf{\mu}, \kappa) = C_{d}(\kappa) \exp \left(\kappa \mathbf{\mu}^\mathrm{T} \mathbf{x} \right)$, where $C_d(\kappa)$ is a normalization constant, $\mathbf{\mu}$ is a unit vector representing the mean direction of the distribution, and $\kappa$ is the concentration parameter that controls the concentration of the distribution around the mean direction. 

The mean shift algorithm is an iterative procedure for finding the maxima of a distribution given a set of sampling points. In each mean shift iteration, the cluster centers will be updated with a shift towards the mean vector of its neighborhood, a set of sampling points within a certain window. The cluster centers will reach the positions with highest local densities once converged. Kobayashi and Otsu \cite{Kobayashi2010} proposed the use of the vMF distribution as a window function for sampling points on the hypersphere. In each mean shift iteration, each clustering center is updated by 
\begin{equation} \label{msupdate}
\mu_{t+1} = \frac{\sum^N_i \mathbf{x}_i \exp \left(\kappa \mu_t^\mathrm{T} \mathbf{x}_i \right)}{\| \sum^N_i \mathbf{x}_i \exp \left(\kappa \mu_t^\mathrm{T} \mathbf{x}_i \right) \|},
\end{equation}
where $\mathbf{x}_i, i=1, \ldots, N$ are sampling points from a dataset with size $N$. In this work, we adapt the concept of vMF mean shift clustering in our proposed mean shift mask transformer, which simulates the mean shift clustering procedure using a transformer decoder.
\section{Method}



In this section, we introduce the Mean Shift Mask Transformer (MSMFormer) as shown in Fig. \ref{fig:model}.
We utilize transformer decoders to convert object queries into object masks. By predicting $N$ binary masks, a mask transformer architecture groups pixels into $N$ segments. An inference pipeline in a mask transformer framework has three main steps. The first uses a backbone and an optional pixel decoder to generate feature maps $F$. The backbone extracts image features, which are then upsampled to higher-resolution features by the pixel decoder. Next, a transformer decoder that has been trained with a set of prediction losses produces mask embeddings $C$ by updating object queries with the feature maps $F$. Lastly, binary mask predictions are generated from $\operatorname{softmax} (C \times F)$. We use the pretrained Unseen Clustering Network (UCN) \cite{xiang2020learning} to obtain robust RGB-D pixel embeddings since the backbone has been trained with a metric learning loss function~\cite{xiang2020learning, de2017semantic, xie2019object}.



\vspace{-2mm}
\subsection{Relationship between Cross-Attention and vMF Mean Shift Clustering}


The key problem with the mask transformer-based segmentation frameworks is how to transform the object queries, i.e., some randomly initialized learnable embeddings, into informative mask embedding vectors.  The design of our transformer decoder is inspired by the vMF mean shift clustering. To explain the development of our decoder layer, we begin by describing the similarity of the typical scaled dot-product attention~\cite{vaswani2017attention} and one center update step in the vMF mean shift clustering, which motivates the hypersphere attention that we proposed. The cross-attention is subsequently modified into the attention version of one iteration in the clustering process.

\textbf{Scaled Dot-Product Attention} \cite{vaswani2017attention}. The scaled dot-product attention is defined as:
\vspace{-4mm}
\begin{equation} \label{attention}
\operatorname{Attention}(\mathbf{Q}, \mathbf{K}, \mathbf{V})  = \operatorname{softmax}\left(\frac{\mathbf{Q} \mathbf{K}^\mathrm{T}}{\sqrt{d_k}}\right) \mathbf{V}    = \frac{\exp{\left(\frac{\mathbf{Q} \mathbf{K}^\mathrm{T}}{\sqrt{d_k}}\right)}}{C_n} \mathbf{V},
\end{equation}
where $C_n$ stands for a normalization term. $\mathbf{Q} \in \mathbb{R}^{N \times D_k}$ denotes the $N$ $D_k$-dim query vectors. $\mathbf{K} \in \mathbb{R}^{M \times D_k}$, $\mathbf{V} \in \mathbb{R}^{M \times D_v}$ are $M$ $D_k$-dim key vectors and $M$ $D_v$-dim value vectors, respectively. This function calculates the affinity between queries $\mathbf{Q}$ and keys $\mathbf{K}$ as the weights assigned to values $\mathbf{V}$. Next, it outputs the weighted sum of the values. The \textit{softmax} function decomposes into an exponential function in the numerator and a normalization term in the denominator. 

Comparing Eq. \eqref{msupdate} and Eq. \eqref{attention}, both exponential functions produce outputs that represent the similarity of two items. These outputs are scaled by a factor ($\kappa$ or $\frac{1}{\sqrt{d_k}}$) and then normalized. Ignoring the subtle differences between the above two equations, the weighted sums of values become the outputs. Since the attention process is similar to the update step of the vMF mean shift clustering, we propose the following attention variant.


\begin{figure}
    \centering
\includegraphics[width=0.55\columnwidth]{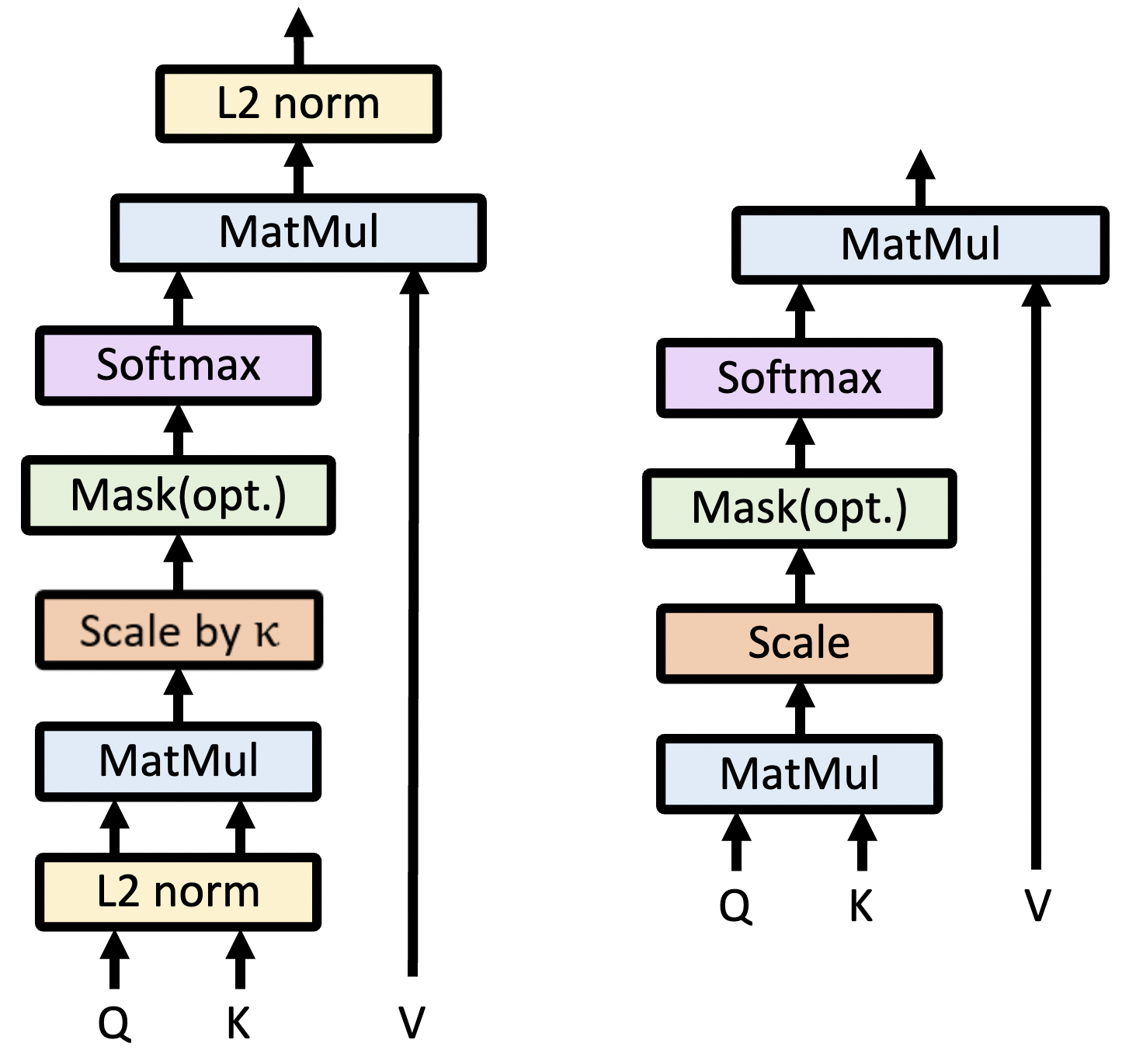}
    \vspace{-4mm}
    \caption{ (left) Our proposed hyper-
sphere attention. (right) The standard
scaled dot-product attention. Mask(Opt.) stands for optional attention masks.}
    \label{fig:attention}
    \vspace{-6mm}
\end{figure}

Our \textbf{Hypersphere Attention} (HSAtten) is defined as:
\begin{equation} \label{HSAtten}
\operatorname{HSAtten}(\mathbf{Q}, \mathbf{K}, \mathbf{V})=g({\operatorname{softmax}}\left(\kappa g(\mathbf{Q})\ g(\mathbf{K})^{\mathrm{T}}\right) \mathbf{V}),
\end{equation}
where $\kappa$ is a positive scalar hyperparameter. $g(\mathbf{x})=\frac{\mathbf{x}}{\lVert \mathbf{x} \rVert}$ is used to $\ell_2$ normalize these vectors into unit vectors. As shown in Fig. \ref{fig:attention}, we first compute the dot products of the normalized query vectors and key vectors as the cosine similarities between $\mathbf{Q}$ and $\mathbf{K}$, which are then scaled by $\kappa$. Then the weights of values are obtained by $\operatorname{softmax}$. Finally, we normalize the weighted sum of value vectors as the output. We call this attention variant \emph{hypersphere attention} because these $\ell_2$ normalized vectors are on a hypersphere. Fig. \ref{fig:attention} demonstrates the difference between our hypersphere attention and the scaled dot-product attention. 
We replace the scaled dot-product attention with our hypersphere attention to build our mean shift self-attention and cross-attention mechanisms in the next section. 

\subsection{Masked Mean Shift Cross-Attention} 
\label{section:mscross-attention}
Simply using hypersphere attention is not enough to obtain good mask embeddings. Therefore, we encapsulate our attention in transformer-decoder layers. Further, inspired by masked attention \cite{cheng2022masked}, we add attention masks into the hypersphere attention in order to only allow a window of image feature embeddings affecting queries. 

\textbf{Cross-attention}. The cross-attention modules are used to assemble pixel features that belong to the same object and update the corresponding object queries, defined as follows:
\begin{equation} \label{crossAtt}
    \hat{\mathbf{X}}=\mathbf{X}+\operatorname{softmax}\left(\mathbf{Q} \times\mathbf{K}^{\mathrm{T}}\right) \times \mathbf{V},
\end{equation}
where $\mathbf{X} \in \mathbb{R}^{N \times D}$ denotes $N$ object queries with $D$ channels, and $\hat{\mathbf{X}}$ is the updated queries using image features. The key-value pairs $\mathbf{K} \in \mathbb{R}^{H W \times D}, \mathbf{V} \in \mathbb{R}^{H W \times D}$ are produced via a linear transformation of the image features, where $H$ and $W$ are the height and width of the image, respectively. 

\textbf{Mean Shift Cross-attention}. In our case, object queries can be viewed as cluster centers on a hypersphere. By substituting the scaled dot-product attention with our hypersphere attention, we obtain:
\begin{equation} \label{mscross}
    \hat{\mathbf{X}}=\mathbf{X}+g(\operatorname{softmax}\left(\kappa \times g(\mathbf{Q}) \times g(\mathbf{K})^{\mathrm{T}}\right) \times \mathbf{V}),
\end{equation}
where $g(\mathbf{x})=\frac{\mathbf{x}}{\lVert \mathbf{x} \rVert}$ is used to $\ell_2$ normalize embeddings. The above equation employs the full feature maps to update the queries. However, global context in the cross-attention layers leads to the slow convergence of models with transformer decoders \cite{gao2021fast, sun2021rethinking}. Cheng et al. \cite{cheng2022masked} hypothesize that local features are sufficient to update queries and self-attention can collect global context. Similarly, in mean shift clustering, the points contained within a window are sufficient to update a cluster center. Attention masks enable the cross-attention layers to focus on the local regions near each cluster center. Multiple mean shift self-attention layers can make queries unique, which saves us from merging cluster centers in mean shift clustering.

After introducing attention masks, our masked mean shift cross-attention works as follows:
\begin{equation}
    \mathbf{X}_l=\mathbf{X}_{l-1} + g(\operatorname{softmax}\left(\mathcal{M}_{l-1}+\kappa g(\mathbf{Q}_l) g(\mathbf{K}_l)^{\mathrm{T}}\right) \mathbf{V}_l),
\end{equation}
where $\mathbf{Q}_l$, $\mathbf{K}_l$ and $\mathbf{T}_l$ stands for the queries, keys and values in $l$-th Transformer decoder layer. $\mathbf{X}_{l-1}$ denotes the input of the $l$-th decoder layer. The attention mask $\mathcal{M}_{l-1}$ \cite{cheng2022masked} at feature location $(x, y)$ is computed with the mask prediction $\mathbf{M}_{l-1}$ via
\vspace{-2mm}
\begin{equation} \label{attMask}
    \mathcal{M}_{l-1}(x, y)=\left\{\begin{array}{ll}
0 & \text { if } \operatorname{M}_{l-1}(x, y)=1 \\
-\infty & \text { otherwise }
\end{array} .\right.
\end{equation}
The mask prediction of the $(l-1)$-th transformer decoder layer is first resized to the resolution of $\mathbf{K}_l$. It is then binarized to be $\mathbf{M}_{l-1} \in\{0,1\}^{N \times H W}$ using a threshold of $0.5$. $\mathbf{M}_0$ is the binary mask from the prediction of the raw queries which has not been processed by the transformer decoder.
\vspace{-2mm}
\subsection{Mean Shift Decoder Layer} 
\vspace{-1mm}
\label{msdecoderlayer}

\begin{figure}
    \centering
\vspace{-6mm}
\includegraphics[width=0.55\columnwidth]{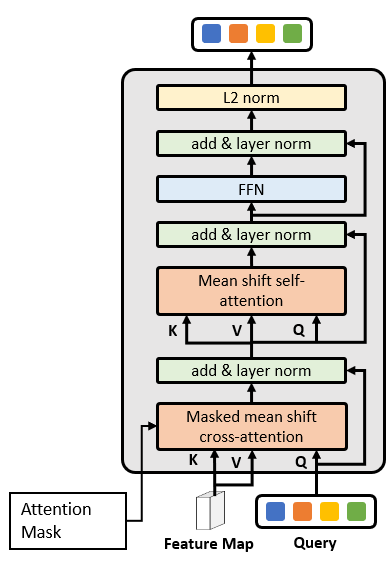}
\vspace{-3mm}
\caption{Our mean shift decoder layer. We have \textit{L2 norm} after FFN.}
\label{fig:decoder}
    \vspace{-6mm}
\end{figure}

As each iteration of vMF mean shift clustering outputs unit vectors. To obtain unit vectors in our framework, we impose $\ell_2$ normalization on the output of the Feed-Forward Network (FFN) in each decoder layer. As shown in Fig. \ref{fig:decoder}, our mean shift decoder layer (MS decoder layer) consists of three components: mean shift cross-attention, mean shift self-attention, and FFN with $\ell_2$ normalization. In this way, we simulate one iteration of the clustering with one mean shift decoder layer. As shown in Fig. \ref{fig:model}, the mean shift decoder consists of multiple MS decoder layers to replace the mean shift clustering process. Built on the backbones and MS decoder, we propose the Mean Shift Mask Transformer (MSMFormer) for segmentation tasks (Fig. \ref{fig:model}). The backbones generate pixel embeddings to guide object queries in the MS decoder. After the queries find their destinations, the queries can be viewed as the final mask embeddings. They are then multiplied with the pixel embeddings to generate pixel similarities. The pixels with positive similarities form object masks \cite{cheng2021per, cheng2022masked}.

\begin{table*}
\vspace{-6mm}
\caption{Unseen Object Instance Segmentation results on OCID and OSD. $+$: the result using a second-stage network, i.e., after zoom-in refinement. *: models trained with the UOAIS-Sim dataset in~\cite{back2022unseen}. \#: models with amodal perception. }\label{allResults}
\vspace{-2mm}
\centering
\scalebox{0.9}{
\begin{tabular}{|l|l|lllllll|lllllll|}
\hline
\multirow{3}{*}{Method}                                  & \multirow{3}{*}{Input} & \multicolumn{7}{c|}{OCID (2390 images)}                                                                                                                                                 & \multicolumn{7}{c|}{OSD (111 images)}                                                                                                                                                   \\ \cline{3-16} 
                                                         &                        & \multicolumn{3}{c|}{Overlap}                                                       & \multicolumn{3}{c|}{Boundary}                                                      &               & \multicolumn{3}{c|}{Overlap}                                                       & \multicolumn{3}{c|}{Boundary}                                                      &               \\
                                                         &                        & \multicolumn{1}{c}{P} & \multicolumn{1}{c}{R} & \multicolumn{1}{c|}{F}             & \multicolumn{1}{c}{P} & \multicolumn{1}{c}{R} & \multicolumn{1}{c|}{F}             & \%75          & \multicolumn{1}{c}{P} & \multicolumn{1}{c}{R} & \multicolumn{1}{c|}{F}             & \multicolumn{1}{c}{P} & \multicolumn{1}{c}{R} & \multicolumn{1}{c|}{F}             & \%75          \\ \hline
MRCNN~\cite{he2017mask}                                                & RGB                    & \textbf{77.6}         & 67.0                  & \multicolumn{1}{l|}{67.2}          & \textbf{65.5}         & 53.9                  & \multicolumn{1}{l|}{54.6}          & 55.8          & 64.2         & 61.3                  & \multicolumn{1}{l|}{62.5}          & 50.2                  & 40.2                  & \multicolumn{1}{l|}{44.0}          & 31.9          \\
UCN \cite{xiang2020learning}                                                      & RGB                    & 54.8                  & 76.0         & \multicolumn{1}{l|}{59.4}          & 34.5                  & 45.0                  & \multicolumn{1}{l|}{36.5}          & 48.0          & 57.2                  & 73.8         & \multicolumn{1}{l|}{63.3}          & 34.7                  & 50.0                  & \multicolumn{1}{l|}{39.1}          & 52.5          \\
UCN+ \cite{xiang2020learning}                                                     & RGB                    & 59.1                  & 74.0                  & \multicolumn{1}{l|}{61.1}          & 40.8                  & 55.0                  & \multicolumn{1}{l|}{43.8}          & 58.2 & 59.1                  & 71.7                  & \multicolumn{1}{l|}{63.8}          & 34.3                  & 53.3         & \multicolumn{1}{l|}{39.5}          & 52.6 \\
Mask2Former \cite{cheng2022masked}                                           & RGB                    & 67.2                  & 73.1                  & \multicolumn{1}{l|}{67.1}          & 55.9                  & 58.1         & \multicolumn{1}{l|}{54.5}          & 54.3          & 60.6                  & 60.2                  & \multicolumn{1}{l|}{59.5}          & 48.2                  & 41.7                  & \multicolumn{1}{l|}{43.3}          & 32.4          \\

UOAIS-Net \cite{back2022unseen}*                                         & RGB \#                   & 66.5                           & 83.1                  & \multicolumn{1}{l|}{67.9}          & 62.1                           & 70.2                           & \multicolumn{1}{l|}{62.3}          & 73.1                           & \textbf{84.2}                  & \textbf{83.7}                  & \multicolumn{1}{l|}{\textbf{83.8}} & \textbf{72.2}                  & \textbf{72.8}                  & \multicolumn{1}{l|}{\textbf{72.1}} & \textbf{76.7}                  \\

MSMFormer* (Ours)                                  & RGB                    & 70.2                  & \textbf{84.4}                           & \multicolumn{1}{l|}{\textbf{70.5}} & 64.5                  & \textbf{74.9}                  & \multicolumn{1}{l|}{\textbf{64.9}} & \textbf{75.3}                  & 59.3                           & 82.0                           & \multicolumn{1}{l|}{67.9}          & 42.9                           & 72.0                           & \multicolumn{1}{l|}{52.4}          & 72.4                           \\ 

MSMFormer (Ours)                                             & RGB                    & 72.9                  & 68.3                  & \multicolumn{1}{l|}{67.7} & 60.5                  & 56.3                  & \multicolumn{1}{l|}{55.8} & 52.9          & 63.4                  & 64.7                  & \multicolumn{1}{l|}{63.6} & 48.6                  & 47.4                  & \multicolumn{1}{l|}{47.0} & 40.2          \\
MSMFormer+ (Ours)                                          & RGB                    & 73.9                  & 67.1                  & \multicolumn{1}{l|}{66.3}          & 64.6                  & 52.9                  & \multicolumn{1}{l|}{54.8}          & 52.8          & 63.9                  & 63.7                  & \multicolumn{1}{l|}{62.7}          & 51.6         & 45.3                  & \multicolumn{1}{l|}{47.0} & 41.1          \\ \hline

MRCNN~\cite{he2017mask}    & Depth                  & 85.3                  & 85.6                  & \multicolumn{1}{l|}{84.7}          & 83.2                  & 76.6                  & \multicolumn{1}{l|}{78.8}          & 72.7          & 77.8                  & 85.1                  & \multicolumn{1}{l|}{80.6}          & 52.5                  & 57.9                  & \multicolumn{1}{l|}{54.6}          & 77.6          \\
UOIS-Net-2D \cite{xie2020best}          & Depth                  & 88.3                  & 78.9                  & \multicolumn{1}{l|}{81.7}          & 82.0                  & 65.9                  & \multicolumn{1}{l|}{71.4}          & 69.1          & 80.7                  & 80.5                  & \multicolumn{1}{l|}{79.9}          & 66.0                  & 67.1                  & \multicolumn{1}{l|}{65.6}          & 71.9          \\
UOIS-Net-3D \cite{xie2021unseen}       & Depth                  & 86.5                  & 86.6                  & \multicolumn{1}{l|}{86.4}          & 80.0                  & 73.4                  & \multicolumn{1}{l|}{76.2}          & 77.2          & 85.7                  & 82.5                  & \multicolumn{1}{l|}{83.3}          & \textbf{75.7}         & 68.9                  & \multicolumn{1}{l|}{71.2}          & 73.8          \\

UOAIS-Net \cite{back2022unseen}*   & Depth \#                 & 89.9                           & 90.9                           & \multicolumn{1}{l|}{89.8} & 86.7                           & 84.1                           & \multicolumn{1}{l|}{84.7} & 87.1                           & 84.9                           & 86.4                           & \multicolumn{1}{l|}{85.5} & 68.2                           & 66.2                           & \multicolumn{1}{l|}{66.9} & 80.8                           \\

UOAIS-Net \cite{back2022unseen}*   & RGBD \#                  & 70.7                           & 86.7                  & \multicolumn{1}{l|}{71.9}          & 68.2                           & 78.5                  & \multicolumn{1}{l|}{68.8}          & 78.7                  & 85.3                  & 85.4                           & \multicolumn{1}{l|}{85.2} & 72.7                  & \textbf{74.3}                  & \multicolumn{1}{l|}{\textbf{73.1}} & 79.1                          
                           \\

UCN \cite{xiang2020learning}          & RGBD                   & 86.0                  & 92.3                  & \multicolumn{1}{l|}{88.5}          & 80.4                  & 78.3                  & \multicolumn{1}{l|}{78.8}          & 82.2          & 84.3                  & \textbf{88.3}         & \multicolumn{1}{l|}{86.2}          & 67.5                  & 67.5                  & \multicolumn{1}{l|}{67.1}          & 79.3          \\

UCN+ \cite{xiang2020learning}         & RGBD                   & 91.6                  & \textbf{92.5}         & \multicolumn{1}{l|}{\textbf{91.6}} & 86.5                  & \textbf{87.1}         & \multicolumn{1}{l|}{86.1}          & \textbf{89.3} & \textbf{87.4}         & 87.4                  & \multicolumn{1}{l|}{\textbf{87.4}} & 69.1                  & 70.8                  & \multicolumn{1}{l|}{69.4}          & \textbf{83.2} \\

Mask2Former \cite{cheng2022masked} & RGBD                   & 78.6                  & 82.8                  & \multicolumn{1}{l|}{79.5}          & 69.3                  & 76.2                  & \multicolumn{1}{l|}{71.1}          & 69.3          & 75.6                  & 79.2                  & \multicolumn{1}{l|}{77.3}          & 54.1                  & 64.0                  & \multicolumn{1}{l|}{58.0}          & 65.2          \\ 

MSMFormer* (Ours)                                  & RGBD                   & 78.2                  & 83.8                           & \multicolumn{1}{l|}{76.6} & 69.1                  & 78.4                           & \multicolumn{1}{l|}{69.6} & 75.3                           & 80.6                           & 86.9                  & \multicolumn{1}{l|}{83.1}          & 65.9                           & 72.9                           & \multicolumn{1}{l|}{68.7}          & 80.2                  \\

MSMFormer (Ours)                                         & RGBD                   & 88.4                  & 90.2                  & \multicolumn{1}{l|}{88.5}          & 84.7                  & 83.1                  & \multicolumn{1}{l|}{83.0}          & 80.3          & 79.5                  & 86.4                  & \multicolumn{1}{l|}{82.8}          & 53.5                  & 71.0                  & \multicolumn{1}{l|}{60.6}          & 79.4          \\

MSMFormer+ (Ours)                                        & RGBD                   & \textbf{92.5}         & 91.0                  & \multicolumn{1}{l|}{91.5}          & \textbf{89.4}         & 85.9                  & \multicolumn{1}{l|}{\textbf{87.3}} & 86.0          & 87.1                  & 86.1                  & \multicolumn{1}{l|}{86.4}          & 69.0                  & 68.6                  & \multicolumn{1}{l|}{68.4}          & 80.4          \\ \hline

\end{tabular}
}
\end{table*}

\section{Experiments}

\subsection{Implementation Details}
\vspace{-1mm}
\textbf{Backbones}. For RGB-D input, we choose two pretrained 34-layer, stride-8 ResNet (ResNet34-8s) from \cite{xiang2020learning} as the backbone for RGB images and depth images, which are pre-trained on the  Tabletop Object Dataset~\cite{xie2019object}. With only RGB input, we choose a pretrained ResNet-50 backbone from Detectron2~\cite{wu2019detectron2} and the pixel decoder from Mask2Former~\cite{cheng2022masked} to compute the feature maps, which achieves better performance than the ResNet-34 backbone in~\cite{xiang2020learning}.

\begin{figure*}[ht]
    \centering
    \includegraphics[width=1.8\columnwidth]{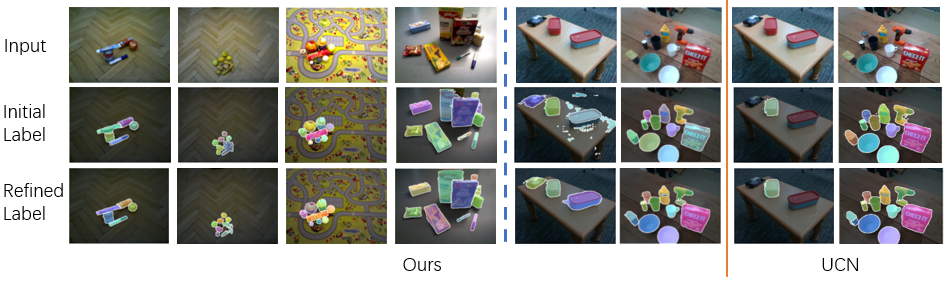}
    \vspace{-5mm}
    \caption{\small Examples of the two-stage predictions. The first six columns show our method's predictions, while the last two columns display UCN's predictions~\cite{xiang2020learning}. Given the same inputs, in the fifth column, our method segments all the objects whereas UCN only finds one of them. In the last image, UCN over-segments the bottom-left can during zoom-in refinement while ours does not.}
    \label{fig:refine}
    \vspace{-6mm}
\end{figure*}

\textbf{Loss Functions}. We adopt the same loss function as Mask2Former \cite{cheng2022masked}. The mask loss consists of the binary cross-entropy loss and the dice loss \cite{milletari2016v}: $\mathcal{L}_{\text {mask }}=\lambda_{\text {ce }} \mathcal{L}_{\text {ce }}+\lambda_{\text {dice }} \mathcal{L}_{\text {dice }}$, where $\lambda_{\text {ce }}$  and $\lambda_{\text {dice }}$ are both set as $5.0$. The final loss is the weighted sum of mask loss and classification loss: $\mathcal{L}_{\text {final }}=\mathcal{L}_{\text {mask }}+\lambda_{\text {cls }} \mathcal{L}_{\text {cls }}$, where $\lambda_{\text {cls }}$ is $2.0$ for predictions that can match with a ground truth object and $0.1$ for predictions that cannot match any ground truth object. We use the AdamW optimizer \cite{loshchilov2017decoupled} with batch size 4 and learning rate 1e-4. We train the model for one epoch on the Tabletop Object Dataset from \cite{xie2020best} that contains 280,000 images. 

\textbf{Post-processing}. We adopt the post-processing from Mask2Former~\cite{cheng2022masked} for instance segmentation. Each mask prediction has a class confidence score. Since the network always outputs 100 mask predictions, we only consider those with a score of 0.7 or higher as reliable masks. 

\textbf{Pixel Confidence Map}. We preserve positive pixel similarities that belong to object masks via the ReLU function. Then we normalize the similarities by dividing them by their maximum value, scaling them from 0 to 1. This process yields pixel confidence heatmaps, which serves as a tool for identifying contour uncertainty~\cite{xie2022rice}.

\textbf{Second Stage as Zoom-in Refinement}. The segmentation accuracy can be improved by using a two-stage clustering process~\cite{xiang2020learning}. Each segmentation mask from the first stage is fed into another MSMFormer for object segmentation again. In this way, nearby objects or stacked objects can be separated using the second-stage segmentation network.

\textbf{Inference Time.} Given a $480 \times 640$ RGB-D image, the running time of MSMFormer is on average 0.278s for the 1st stage and 0.072s per ROI in the 2nd stage on an NVIDIA A100 GPU. In contrast, the reported running time of UCN~\cite{xiang2020learning} is 0.2s for the 1st stage and 0.05s per object in the 2nd stage on a TITAN XP GPU. 

\subsection{Evaluation Results}

\textbf{Datasets}. We evaluate the performance on the Object Clutter Indoor Dataset (OCID) \cite{suchi2019easylabel} and the Object Segmentation Database (OSD) \cite{richtsfeld2012segmentation} for UOIS in tabletop scenes. The OCID dataset contains 2,390 RGB-D images, with up to 20 objects and an average of 7.5 objects per image. The OSD dataset contains 111 RGB-D images, with up to 15 objects and an average of 3.3 objects per image. 

\textbf{Evaluation Metrics}. We evaluate the object segmentation performance using precision, recall and F-measure according to \cite{xie2020best, xiang2020learning}. For these three metrics, we first compute the values between all pairs of predicted objects and ground truth objects. Then we use the Hungarian method with pairwise F-measure to match predictions with ground truth. Based on this matching, the final precision, recall and F-measure are obtained by $P=\frac{\sum_i\left|c_i \cap g\left(c_i\right)\right|}{\sum_i\left|c_i\right|}, R=\frac{\sum_i\left|c_i \cap g\left(c_i\right)\right|}{\sum_j\left|g_j\right|}, F=\frac{2 P R}{P+R}$, where $c_i$ denotes the segmentation of predicted object $i, g\left(c_i\right)$ denotes the segmentation of the matched ground truth object of $c_i$, and $g_j$ is the segmentation for ground truth object $j$. Overlap $\mathrm{P} / \mathrm{R} / \mathrm{F}$ denote the aforementioned three metrics since the true positives can be viewed as the overlap segmentation of the whole object. In addition, Boundary $\mathrm{P} / \mathrm{R} / \mathrm{F}$ are used to evaluate how sharp the predicted boundary matches against the ground truth boundary, where the true positives are counted by the pixel overlap of the two boundaries. Moreover, Overlap F-measure $75 \%$ is the percentage of segmented objects with Overlap F-measure $\geq 75 \%$~\cite{ochs2013segmentation}. 

\begin{table*}
\vspace{-3mm}
\caption{Two kinds of decoder layers results. \textit{Masked} denotes the Transformer decoder layer with masked attention \cite{cheng2022masked}, whereas \textit{MS} is our MS decoder layer. $+$: the two-stage performance. }\label{decoders}
\vspace{-3mm}
\centering
\scalebox{1.0}{
\begin{tabular}{|l|lllllll|lllllll|}
\hline
\multirow{3}{*}{Method}              & \multicolumn{7}{c|}{OCID (2390 images)}                                                                                                                  & \multicolumn{7}{c|}{OSD (111 images)}                                                                                                                    \\ \cline{2-15} 
                                     & \multicolumn{3}{c|}{Overlap}                                       & \multicolumn{3}{c|}{Boundary}                                      &               & \multicolumn{3}{c|}{Overlap}                                       & \multicolumn{3}{c|}{Boundary}                                      &               \\
                                     & \multicolumn{1}{c}{P}    & \multicolumn{1}{c}{R}    & \multicolumn{1}{c|}{F}    & \multicolumn{1}{c}{P}    & \multicolumn{1}{c}{R}    & \multicolumn{1}{c|}{F}             & \%75          & \multicolumn{1}{c}{P}    & \multicolumn{1}{c}{R}    & \multicolumn{1}{c|}{F}    & \multicolumn{1}{c}{P}    & \multicolumn{1}{c}{R}    & \multicolumn{1}{c|}{F}             & \%75          \\ \hline
Masked  & 88.4          & 90.0          & \multicolumn{1}{l|}{88.2}          & 85.4          & 82.4          & \multicolumn{1}{l|}{83.0}          & 78.8          & 72.4          & 80.5          & \multicolumn{1}{l|}{76.2}          & 45.6          & 63.4          & \multicolumn{1}{l|}{52.5}          & 65.3          \\
Masked+ & 90.3          & 89.2          & \multicolumn{1}{l|}{89.4}          & 87.7          & \textbf{89.2} & \multicolumn{1}{l|}{85.0}          & 81.5          & 80.1          & 79.6          & \multicolumn{1}{l|}{79.7}          & 64.1          & 59.8          & \multicolumn{1}{l|}{61.4}          & 66.5          \\ \hline
MS (Ours)             & 88.4          & 90.2          & \multicolumn{1}{l|}{88.5}          & 84.7          & 83.1          & \multicolumn{1}{l|}{83.0}          & 80.3          & 79.5          & \textbf{86.4} & \multicolumn{1}{l|}{82.8}          & 53.5          & \textbf{71.0} & \multicolumn{1}{l|}{60.6}          & 79.4          \\
MS+ (Ours)   & \textbf{92.5} & \textbf{91.0} & \multicolumn{1}{l|}{\textbf{91.5}} & \textbf{89.4} & 85.9          & \multicolumn{1}{l|}{\textbf{87.3}} & \textbf{86.0} & \textbf{87.1} & 86.1          & \multicolumn{1}{l|}{\textbf{86.4}} & \textbf{69.0} & 68.6          & \multicolumn{1}{l|}{\textbf{68.4}} & \textbf{80.4} \\ \hline
\end{tabular}
}
\vspace{-2mm}
\end{table*}

\textbf{Comparison to other Methods}. In Table \ref{allResults}, we compare our model with existing methods on OCID and OSD. For OCID, our RGB-D model achieves a new state-of-the-art of 87.3 Boundary F-measure. The first-stage model already reaches 83.0 Boundary F-measure which demonstrates the advantage of MSMFormer on accurately segmenting objects. With only RGB images, our model also outperforms other methods on the OCID dataset.

For OSD, the UOAIS-Net~\cite{back2022unseen} achieves the best performance by using a new synthetic dataset UOAIS-Sim for training, 
which empolys Hierarchical Occlusion Modeling trained with the amodal and visible masks. We found that training MSMFormer only on the visible masks of this new synthetic dataset improves our results on OCID but not on OSD. The main reason is that OSD contains images with heavy occlusions. The UOAIS-Sim dataset provides more training images in these scenarios.

Synthetic training datasets impose limitations on our model's performance. The UOAIS-Sim dataset with superior quality RGB images in comparison to the TableTop dataset, contributes to the enhancement of MSMFormer's performance. Our model demonstrates robust scalability when exposed to improved datasets.



\textbf{Effect of Second-stage Model}. For RGB-D input, the results of the model with or without the second-stage refinement are presented in the final two rows of Table \ref{allResults}. The second-stage model improves most of the metrics in both data sets. For RGB input, the first-stage model makes incorrect predictions on backgrounds such as walls and colorful blankets. The second stage model does not remove these false alarms. Therefore, it cannot enhance the overall segmentation performance. Fig. \ref{fig:refine} shows some segmentation results before and after refinement using RGB-D input. In these examples, the boundaries are refined and more consistent, and some merged objects are separated correctly. 



\vspace{-2mm}
\subsection{Ablation Studies}
\vspace{-2mm}

\textbf{Our Mean Shift Decoder vs. Standard Scaled Dot-product Attention Decoder}. We train Mask2Former \cite{cheng2022masked} using RGB-D images with ResNet50 backbones on the Tabletop Object Dataset. As shown in Table~\ref{allResults}, our MSMFormer significantly outperforms Mask2Former on both OCID and OSD. The main reason is that Mask2Former overfits the synthetic training set and does not generalize well to real images. It cannot benefit from the pretrained backbones trained with the metric learning loss from UCN \cite{xiang2020learning}. Therefore, we use the same pretrained network from UCN to compare our MS decoder layer and the Masked-attention Transformer decoder \cite{cheng2022masked}. They are trained for one epoch on the Tabletop Object dataset. In Table \ref{decoders}, our MS decoder layer outperforms the Masked-attention decoder in most metrics on both datasets. 

\begin{table}
  \centering
  \vspace{-3mm}
\caption{Ablation study on different numbers of MS decoder layers.}
\vspace{-2mm}
\scalebox{0.9}{
\begin{tabular}{|c|c|ccccccc|}
\hline
\multirow{3}{*}{Stage}       & \multirow{3}{*}{\begin{tabular}[c]{@{}c@{}}\#layers\end{tabular}} & \multicolumn{7}{c|}{OCID(2390 images)}                                                                                                                  \\ \cline{3-9} 
                              &                                                                                    & \multicolumn{3}{c|}{Overlap}                                       & \multicolumn{3}{c|}{Boundary}                                      &               \\
                              &                                                                                    & P             & R             & \multicolumn{1}{c|}{F}             & P             & R             & \multicolumn{1}{c|}{F}             & \%75          \\ \hline
\multirow{3}{*}{1st}  & 4                                                                                  & 74.4          & 83.2          & \multicolumn{1}{c|}{78.4}          & 45.8          & 67.0          & \multicolumn{1}{c|}{54.0}          & 72.9          \\
                              & 6                                                                                  & \textbf{88.4} & \textbf{90.2} & \multicolumn{1}{c|}{\textbf{88.5}} & 84.7          & \textbf{83.1} & \multicolumn{1}{c|}{\textbf{83.0}} & \textbf{80.3} \\
                              & 8                                                                                  & \textbf{88.4} & 89.2          & \multicolumn{1}{c|}{88.1}          & \textbf{84.8} & 82.8          & \multicolumn{1}{c|}{82.9}          & 80.0          \\ \hline
\multirow{3}{*}{2nd} & 6                                                                                  & 91.8          & 89.6          & \multicolumn{1}{c|}{90.5}          & 88.3          & 83.8          & \multicolumn{1}{c|}{85.6}          & 83.0          \\
                              & 8                                                                                  & \textbf{92.5} & \textbf{91.0} & \multicolumn{1}{c|}{\textbf{91.5}} & \textbf{89.4} & \textbf{85.9} & \multicolumn{1}{c|}{\textbf{87.3}} & \textbf{86.0} \\
                              & 10                                                                                 & 91.6          & 89.7          & \multicolumn{1}{c|}{90.3}          & 88.3          & 83.7          & \multicolumn{1}{c|}{85.5}          & 83.2          \\ \hline
\end{tabular}
  }
  \label{2dec}
\vspace{-4mm}
\end{table}

\textbf{Number of Mean Shift Decoder Layers}. The first-stage model processes RGB-D images with size $480\times640$, whereas the second stage processes ROI images with size $224\times224$. The input image size has an impact on the number of MS decoder layers required, which can be viewed as the number of iterations in vMF mean shift clustering. Table \ref{2dec} shows that with more than 6 decoder layers, the network performance is stable. The second-stage segmentation is based on the output of the best first-stage model with 6 MS decoder layers. In Table \ref{2dec}, even though the results of second-stage networks with various numbers of layers are close, the 8-layer model has the best performance.

\begin{figure}
    \centering
\includegraphics[width=0.95\columnwidth]{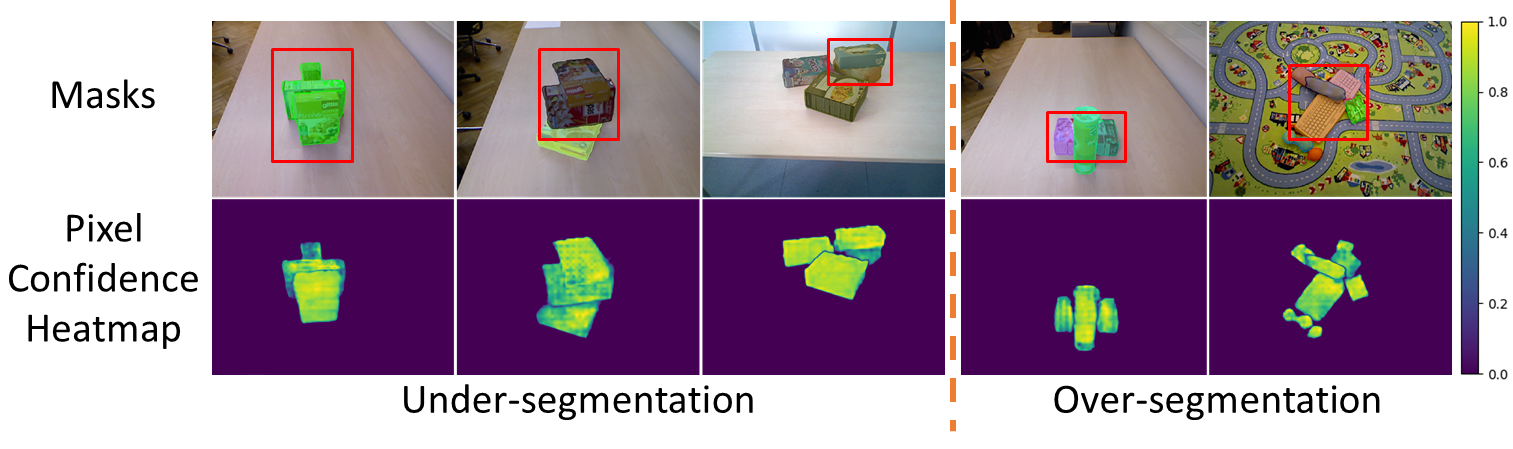}
    \vspace{-4mm}
    \caption{Examples of failure cases and their pixel confidence heatmaps. The contour uncertainty can be found in some under-segmentation cases.}
    \label{fig:failure}
    \vspace{-4mm}
\end{figure}

\textbf{Failure Cases and Limitations}. As shown in Fig. \ref{fig:failure}, most failures are under-segmentation and over-segmentation. In RGB-D images, as some objects are adjacent or stacked together, their embeddings are still close in feature space. These objects are hard to separate even with the second-stage refinement. In Fig. \ref{fig:failure}, for example, the first image has three boxes next to each other, which makes the network struggle to separate them. However, the pixel confidence heatmap can still provide some uncertainty measurement of the under-segmented masks. When an object can be covered by other objects, over-segmentation occurs since the model is misled by the view. For instance, in the last image, the keyboard under other objects is over-segmented. MSMFormer is currently limited by these challenging scenarios.

These issues may be resolved in future work by generating more synthetic training images of these difficult scenarios.

\textbf{Testing with a Real Robot.} We have tested the performance of MSMFormer on a Fetch mobile manipulator. RGB-D images from the Fetch head camera are fed into the network to segment objects in tabletop scenes. Once objects are segmented, we have enabled the robot to pick up objects by applying the Contact-GraspNet~\cite{sundermeyer2021contact} for grasp planning and MoveIt~\cite{chitta2012moveit} for motion planning of grasping. Please see the appendix for additional results and the video for the real-world robot grasping demonstration. 
\vspace{-1mm}
\section{Conclusion and Future Work}
\vspace{-1mm}
In this work, we have proposed a novel transformer attention variant, called \emph{hypersphere attention}, motivated by the von Mises-Fisher mean shift clustering algorithm. With this attention, we designed the masked mean shift cross-attention and mean shift self-attention. Next, we introduced the mean shift decoder for differentiable mean shift clustering to build the Mean Shift Mask Transformer (MSMFormer) for object segmentation tasks. In our model, the feature representation and the clustering for object segmentation can be trained jointly in an end-to-end fashion. In this way, our MSMFormer obtains very competitive results on the Object Clutter In-door Dataset (OCID) and the Object Segmentation
Database (OSD) for Unseen Object Instance Segmentation (UOIS). We also demonstrated that our method is effective for UOIS on real-world images.

For future work, we will explore several ways to further improve the unseen object segmentation performance. One idea is to utilize photo-realistic synthetic images for training. Another idea is to leverage a small set of real-world images to fine-tune the segmentation network for domain adaptation. In addition, we plan to study UOIS beyond tabletop scenes such as bin-picking settings and objects inside cabinets, which will facilitate robot manipulation in these scenarios.

\vspace{-1mm}
\section*{ACKNOWLEDGMENT}
\vspace{-1mm}
This work was supported in part by the DARPA Perceptually-enabled Task Guidance (PTG) Program under contract number HR00112220005.



\bibliographystyle{IEEEtran}
\bibliography{reference}





\end{document}